\documentclass[runningheads]{llncs}

\usepackage[mobile]{eccv}

\usepackage{eccvabbrv}

\usepackage{graphicx}
\usepackage{booktabs}
\usepackage{bm}
\usepackage{color}
\usepackage{pifont}
\usepackage{xcolor}
\usepackage{colortbl}
\usepackage{multirow}
\usepackage{siunitx} % for consistent number formatting
\usepackage{svg}
\usepackage{wrapfig}

\definecolor{rowhighlight}{gray}{0.9}

\usepackage[accsupp]{axessibility}  % Improves PDF readability for those with disabilities.

\usepackage{hyperref}

\usepackage{orcidlink}

\begin{document}

\title{latentSplat: Autoencoding Variational Gaussians for Fast Generalizable 3D Reconstruction
%\vspace{-0.5cm}
} 

\titlerunning{latentSplat: Variational Gaussians}

% TODO OCRID ?
\author{Christopher Wewer\inst{1}\orcidlink{0009-0009-6519-9923}
\and
Kevin Raj\inst{1}\orcidlink{0009-0007-3271-5990}
\and \\
Eddy Ilg\inst{2}\orcidlink{0000-0002-3031-8672}
\and
Bernt Schiele\inst{1}\orcidlink{0000-0001-9683-5237}
\and
Jan Eric Lenssen\inst{1}\orcidlink{0000-0003-4093-9840}
}

\authorrunning{C.~Wewer et al.}

\institute{
\textsuperscript{1}Max Planck Institute for Informatics, Saarland Informatics Campus, Germany\\
\textsuperscript{2}Saarland University, Saarland Informatics Campus, Germany\\
\email{\{cwewer, jlenssen\}@mpi-inf.mpg.de}
}

\maketitle

\vspace{-0.4cm}

\begin{figure}[ht]
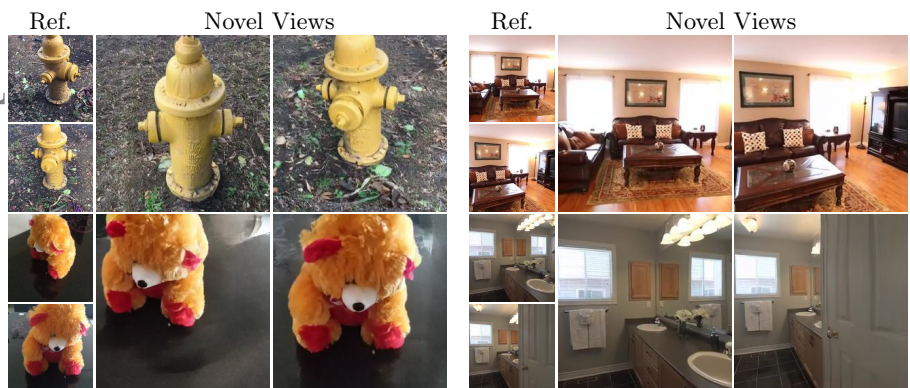

    \centering
    \subfloat{{\includesvg[width = 0.48\textwidth]{co3d_teaser.svg} }}%
    \hspace{0.1cm}
    \subfloat{{\includesvg[width = 0.48\textwidth]{re10k_teaser3.svg} }}%
    %\vspace{-0.2cm}
    \caption{
    \small 
    %\textbf{latentSplat}, a method for fast generalizable 3D reconstruction. We encode two reference views (left) into 3D variational Gaussians. From this representation, we can synthesize novel views (right), generalizing to interpolated and extrapolated views.
    We present \textbf{latentSplat}, a method for scalable generalizable 3D reconstruction from two reference views (left). We autoencode the views into a 3D latent representation consisting of variational feature Gaussians. From this representation, we can perform fast novel view synthesis (right), generalizing to interpolated and extrapolated views.
    }
    \label{fig:teaser}
    \vspace{-0.9cm}
\end{figure}

\begin{abstract}
We present latentSplat, a method to predict semantic Gaussians in a 3D latent space that can be splatted and decoded by a light-weight generative 2D architecture. 
Existing methods for generalizable 3D reconstruction either do not scale to large scenes and resolutions, or are limited to interpolation of close input views.
%Existing methods for generalizable 3D reconstruction either do not enable fast inference of high resolution novel views due to slow volume rendering, or are limited to interpolation of close input views, even in simpler settings with a single central object, where 360-degree generalization is possible. 
latentSplat combines the strengths of regression-based and generative approaches while being trained purely on readily available real video data.
%In this work, we combine a regression-based approach with a generative model, moving towards both of these capabilities within the same method, trained purely on readily available real video data. 
The core of our method are variational 3D Gaussians, a representation that efficiently encodes varying uncertainty within a latent space consisting of 3D feature Gaussians. 
From these Gaussians, specific instances can be sampled and rendered via efficient splatting and a fast, generative decoder. 
%From these Gaussians, specific instances can be sampled and rendered via efficient Gaussian splatting and a fast, generative decoder network. 
We show that latentSplat outperforms previous works in reconstruction quality and generalization, while being fast and scalable to high-resolution data. 
Our project website\footnote{\href{https://geometric-rl.mpi-inf.mpg.de/latentsplat/}{geometric-rl.mpi-inf.mpg.de/latentsplat/}} provides additional results, videos, and code.

%\vspace{-0.1cm}

\end{abstract}

\section{Introduction}
\label{sec:intro}
Performing 3D reconstruction from a single or a few images is a longstanding goal in computer vision, which went through many iterations of advancements, most recently driven by new techniques in the areas of generalizable radiance fields~\cite{yu2021pixelnerf, chen2021mvsnerf} and foundational diffusion models~\cite{glide, dalle2, imagen, stable_diffusion, sdxl}. At the core, the task is to find an optimal 3D representation that fits a set of observations, a task that is highly underconstrained as there are usually an infinite amount of valid reconstructions that satisfy the given observations. Thus, a strong prior is needed to find a fitting solution - usually modeled by deep neural networks trained on a large amount of data. When designing methods to learn data priors for 3D reconstruction, efficiency is a crucial aspect to allow training on large datasets required for generalization. In this work, we present a method that is highly efficient, scales to a large amount of data and can be trained on real video data without 3D supervision. Real video data is readily available in vast quantities and a promising data type for large 3D models. 

Recently, there have been two lines of solutions for the given task of generalizable reconstruction: \emph{regression-based} approaches and \emph{generative} approaches. Methods based on regression, such as pixelNeRF~\cite{yu2021pixelnerf} or pixelSplat~\cite{charatan23pixelsplat}, are usually efficient but are only trained to predict the mean of all possible solutions. While they often succeed in predicting high-quality reconstructions of regions that strongly correlate with the input observations, they struggle in regions of high uncertainty, collapsing to blurry reconstructions that lack high frequency details or fail entirely in unseen areas of larger scenes.

The high ambiguity of solutions in reconstruction from incomplete observations suggests to model it in a probabilistic fashion, that is, obtaining a distribution of possible reconstructions that allows to sample individual solutions. Generative approaches, such as Zero-1-to-3~\cite{liu2023zero1to3} or GeNVS~\cite{chan2023genvs}, follow this principle, allowing to obtain one possible, realistic reconstruction that might contain hallucinated details. 
Here, it is important to note that uncertainty in the case of 3D reconstruction varies heavily depending on 3D location. Some areas of the reconstructed scene are observed directly, maybe even by a lot of views, while others are fully occluded or subject to high ambiguity due to very sparse observation. Thus, a sophisticated reconstruction method that models uncertainty should account for varying amounts of uncertainty in 3D space, and needs a generative model to obtain high-quality reconstructions in uncertain areas. 

Regression-based models have often been used as conditioning for generative models. 
A well-known example are Variational Autoencoders (VAEs)~\cite{kingma2014}. An encoder is used to parameterize a \emph{variational distribution} in latent space, from which we can sample vectors that can be decoded into elements following the data distribution. It is promising to bring this concept efficiently to 3 dimensions, where a regression model estimates the uncertainty for different locations in 3D space individually, providing the desired locality in uncertainty modeling.

In this work, we approach the desired goals by introducing \emph{latentSplat}, a fast method for generalizable 3D reconstruction that combines the strengths of regression-based and generative approaches. As core of latentSplat we introduce \emph{variational 3D Gaussians}, a representation that models uncertainty explicitly by holding distributions of semantic features on predicted locations in 3D space. Variational Gaussians are obtained via an encoder from two images and model varying amounts of uncertainty depending on the location in 3D space. In observed locations, they can provide a regressed solution with low variance, while acknowledging uncertainty in unobserved areas. From a set of variational Gaussians in 3D space, we can sample a specific instance via the reparameterization trick, render it via efficient splatting to arbitrary views, and decode it with an efficient, generative decoder network in pixel space. We show that \emph{latentSplat}:
\begin{itemize}
    \item outperforms recent methods in two-view reconstruction, achieving state-of-the-art quality both quantitatively and qualitatively, especially in challenging cases of wide-spread input views and view extrapolation, 
    \item is fast and efficient in training and rendering, providing a more scalable solution than previous generative methods,
    \item is applicable to both, object-centric (with 360° views) and general scenes,
    \item enables downstream mesh reconstruction via 3D consistent novel views,
    \item is purely trained on real videos, which is a readily available data resource.
\end{itemize}

\section{Related Work}
We revisit recent methods in the area of generalizable novel view synthesis (NVS) and 3D reconstruction. Neural fields~\cite{occupancynet, deepsdf, nerf, deepls, siren} have been the dominant representation to store 3D information for single scenes and scene/object distributions. Lately, more explicit representations live through a renaissance with 3D Gaussian Splatting~\cite{kerbl20233Dgaussians}. Due to the challenging nature of the task, most methods that enable generalization to 360\textdegree~novel views only perform reconstruction on the level of individual objects. Even with the recent rise of large-scale generative models, methods that perform extrapolation of larger scenes are still a rarity, often due to missing scalability. In the following, we distinguish between regression-based approaches in Sec.~\ref{sec:regression} and generative approaches in Sec.~\ref{sec:generative}.

\subsection{Regression-based Generalizable NVS}
\label{sec:regression}
Several regression-based models for 3D reconstruction from a few views have been proposed in recent years. One line of work performs generalization over object categories~\cite{srn, codenerf, visionnerf, fe-nvs, wewer23simnp} only. On scene level, early methods~\cite{yu2021pixelnerf, chibane2021srf, Niemeyer2020diffvol} focus on small-scale setups~\cite{dtu} because of limited capacity or efficiency. Larger scenes require scalable approaches. Image-conditioned neural radiance fields~\cite{neo360} fail in scaling to high resolutions. Image-based rendering methods~\cite{pbgraphics, pbgraphics+, riegler2020fvs, Wiles2020synsin} produce high-quality results in view interpolation but cannot generalize to unseen areas. A related approach is to predict multi-plane images~\cite{Srinivasan2019pushing, tucker2020singleviewmpi, zhang2022video, Zhou2018multiplane}, which is limited to small view-point variations only. %Since generalization is limited to multiple planes in 3D, these methods only allow small view-point variations. 
Multi-view stereo is also a popular way to provide geometry priors for novel view synthesis with deep learning~\cite{chen2021mvsnerf, riegler2020fvs, johari2022geonerf}. Moreover, several alternative representations have been introduced, such as neural rays~\cite{liu2022neuray}, light fields~\cite{sitzmann2021lfns, Suhail2022lighfield}, and patches~\cite{suhail2022patchbased}.

In contrast to all of the above, we provide high-quality 360° reconstructions of object-centric scenes as well as view inter- and extrapolation on large scenes, given only two input views.
Closest to our work are the very recent Splatter Image~\cite{splatterimage} and pixelSplat~\cite{charatan23pixelsplat} following the success of 3D Gaussian Splatting~\cite{kerbl20233Dgaussians} in many domains~\cite{npg, 4dgs, sugar, 2dgs}. In contrast to their purely regression-based approaches, we (1) introduce a semantic feature representation instead of purely explicit Gaussians and (2) model uncertainty explicitly enabling correct generalization to out-of-context views. Thus, we are able to reconstruct full objects in high quality, even if they are only partially observed with two views.

\subsection{Generative Models for NVS}
\label{sec:generative}
Generative approaches succeed in situations with high uncertainty, i.e. when the conditioning is not sufficient to determine the full reconstruction, or if there is no conditioning at all.
A large line of work performs 3D reconstruction of objects with 3D-aware GAN architectures via conditional sampling or inversion~\cite{graf, pigan, disentangled3d, eg3d, meshinversion, pavllo2023shape}. While these methods are able to produce high-quality results, they are not applicable to scenes.
Autoregressive transformers have been shown to be able to synthesize novel views that are consistent to some extent with the past sequence of views, but they fail to successfully leverage explicit 3D biases~\cite{rombach2021geometryfree, ren2022look}.
Following the success of diffusion models~\cite{ddpm, guideddiffusion} for large-scale text-to-image generation~\cite{glide, imagen, dalle2, stable_diffusion, sdxl}, there are many approaches to adapting the same concepts for 3D.
One line of research trains diffusion models directly on 3D representations such as voxel grids~\cite{diffrf, holodiff, holofusion}, triplanes~\cite{triplanediff, ssdnerf}, or point clouds~\cite{schroppel23npcd, pc2, xie23hdm}, and can implement 3D reconstruction by conditioning or guiding the diffusion process with gradients from reconstructing input views.
Similar to 3D-aware GANs, another line~\cite{renderdiff, viewsetdiff, forwarddiff, diffviarender} integrates rendering of a 3D representation into the denoising architecture for image diffusion to achieve 3D generation and reconstruction purely trained on images.
However, both of these approaches do not scale to large scenes due to expensive 3D architectures or slow rendering within the sampling process, respectively.
A third group of works~\cite{3dim, zhou2023sparsefusion, liu2023zero1to3, chan2023genvs, zeronvs, reconfusion} extends 2D diffusion models with pose-conditioning for NVS with the additional benefit of compatibility with pre-trained text-to-image generators~\cite{stable_diffusion} as strong priors.
Closing the loop with regression-based approaches, a particularly effective pose-conditioning proposed by GeNVS~\cite{chan2023genvs} is rendering of pixelNeRF~\cite{yu2021pixelnerf} features.
While scalable to large scenes, these approaches inherit the slow sampling of diffusion models for NVS.
%and, even worse, employ cumbersome iterative optimization to fuse repeatedly generated novel views in a consistent manner at test time to achieve 3D reconstruction.

In contrast, our approach is orders of magnitude faster and scales easily to high resolutions due to the efficient Gaussian representation instead of volume rendering, and a lightweight decoder instead of expensive diffusion sampling.

\begin{figure}[t]
\centering
\includegraphics[width=1.15\textwidth]{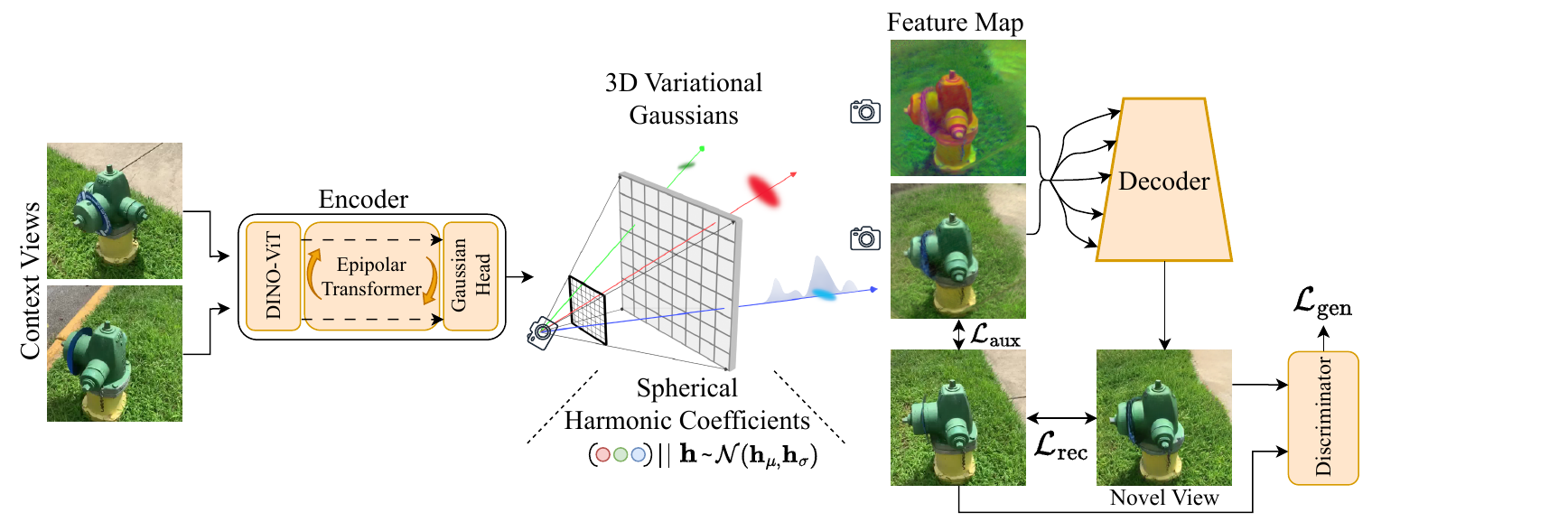}
    \caption{
    %\small 
    %We present latentSplat, a method for scalable generalizable 3D reconstruction from two views. 
    \textbf{latentSplat architecture.}
    The architecture follows an autoencoder structure. \textbf{(Left)} Two input views are encoded into a 3D variational Gaussian representation using an epipolar transformer and a Gaussian sampling head. \textbf{(Center)} Variational Gaussians allow sampling of spherical harmonics feature coefficients that determine a specific instance of semantic Gaussians. \textbf{(Right)} The sampled instance can be rendered efficiently via Gaussian splatting and a light-weight VAE-GAN decoder.}
    \label{fig:method}
\end{figure}

\section{Autoencoding Variational Gaussians}
\label{sec:method}
In this section, we describe our method in detail, beginning with describing our reconstruction task in Sec.~\ref{sec:task}, before introducing the core of our framework, the semantic variational Gaussian representation, in Sec.~\ref{sec:variational_gaussians}. Then, Sec.~\ref{sec:encoder} and Sec.~\ref{sec:decoder} will introduce the encoder and decoder architectures, respectively. Last, training details and loss functions are given in Sec.~\ref{sec:training}.

\subsection{Overview and Assumptions}
\label{sec:task}
We aim to achieve novel view synthesis from two given video frames (reference views) as input. We assume a dataset of videos with camera poses for each frame such that we can build triplets of two reference views and a target view used for training of our model.
As outlined in Fig.~\ref{fig:method}, our method consists of an encoder, encoding a pair of reference views into a 3D latent representation of Gaussians, the variational Gaussians themselves, and a decoder, rendering the Gaussians from arbitrary views.
During training we optimize all parameters to reconstruct the ground-truth target view given the two input images, their camera poses, and the target pose.
Once we trained the model, we can obtain variational Gaussians from two views and render them to synthesize novel views.

\subsection{Variational 3D Gaussians}
\label{sec:variational_gaussians}
At the core of the presented method is a 3D representation that encodes the scene as a set of semantic 3D Gaussians, describing the scene appearance via attached view-dependent feature vectors. In addition, we model uncertainty for each semantic Gaussian individually by storing parameters $\mu$ and $\sigma$ of a normal distribution of spherical harmonic coefficients instead of explicit feature vectors. In total, a scene is represented as a set of $N$ \emph{variational Gaussians} denoted as
\begin{equation}
    \mathcal{G} = \{(\mathbf{x}, \mathbf{S},  \mathbf{R}, o, \mathbf{c}, \mathbf{h}_\mu, \mathbf{h}_\sigma)_i\}_{1\leq i \leq N} \textnormal{,}
\end{equation}
where $\mathbf{x}\in \mathbb{R}^3$ is the three-dimensional location, diagonal matrix $\mathbf{S}\in \mathbb{R}^{3\times3}$ Gaussian scale, $\mathbf{R} \in \mathbb{R}^{3\times3}$ Gaussian orientation, $o \in [0,1]$ the opacity of the Gaussian in 3D space, and $\mathbf{c}\in \mathbb{R}^K$ the spherical harmonics for view-dependent colors. Scale and rotation form the covariance of the 3D Gaussian in space, i.e. $\mathbf{C} = \mathbf{R}\mathbf{S}\mathbf{S}^\top\mathbf{R}^\top$~\cite{kerbl20233Dgaussians}. The variational distributions of Gaussian features are modeled in the coefficient space of spherical harmonics by parameters $\mathbf{h}_\mu, \mathbf{h}_\sigma$ defining normal distributions $\mathcal{N}(\mathbf{h}_\mu, \textnormal{diag}(\mathbf{h}_\sigma))$. The $\mathbf{h}_\sigma$ hold uncertainty information for individual locations in $\mathbb{R}^3$. We still optimize for explicit RGB coefficients $\mathbf{c}$ in addition to feature parameters $\mathbf{h}_\mu \text{ and } \mathbf{h}_\sigma \in \mathbb{R}^4$, as Gaussian shape parameters are best optimized on RGB signals (c.f. Sec.~\ref{sec:decoder}).

\vspace{0.3cm}
\noindent\textbf{Sampling Semantic Gaussians.} 
%\paragraph{Sampling Semantic Gaussians} 
We distinguish between two states of our Gaussian representation, \emph{variational Gaussians} and \emph{semantic Gaussians}. The latter can be obtained from the former by sampling explicit spherical harmonic coefficients via the reparameterization trick for all Gaussians:
\begin{equation}\label{eq:sampling}
    \mathbf{h} = \mathbf{h}_\mu + \epsilon\cdot\mathbf{h}_\sigma \quad \quad \epsilon \sim \mathcal{N}(0, \mathbf{1}) \textnormal{,}
\end{equation}
allowing to backpropagate gradients from the sampled coefficients to the reference view encoder. Intuitively, \emph{variational Gaussians} describe the distribution of all possible 3D reconstructions, conditioned on the given reference views. In contrast, \emph{semantic Gaussians} represent a specific sample from this distribution, allowing consistent multi-view renderings of one possible reconstruction.

\vspace{0.3cm}
\noindent\textbf{Rendering Semantic Gaussians.} 
%\paragraph{Rendering Semantic Gaussians}
A set of semantic Gaussians can be rendered via the efficient Gaussian splatting renderer provided by Kerbl et al.~\cite{kerbl20233Dgaussians}. We extended it to render feature vectors in addition to RGB colors. The spherical harmonic basis is used to decode our per-Gaussian coefficients $\mathbf{h}$ into view-dependent features before splatting them into pixel space. 
The encoder architecture presented in Sec.~\ref{sec:encoder} predicts Gaussians in the field of view of two reference views. Thus, when rendering novel views, there might be regions in the rendered images that do not have any Gaussians, because they are outside of all reference view frustums. In order to provide a plausible reconstruction of these areas, we obtain our feature image $\mathbf{F}$ by sampling from the normal distribution $\mathcal{N}(\mathbf{F}^\textnormal{ren}, 1-\mathbf{O})$ with rendered features $\mathbf{F}^\textnormal{ren}$ and opacity $\mathbf{O}$ via the reparameterization trick: %as a combination of rendered features $\mathbf{F}^\textnormal{ren}$ and pure Gaussian noise, linearly interpolated based on rendered opacity $\mathbf{O}$:
\begin{equation}\label{eq:rendering}
\mathbf{F} =  \mathbf{F}^\textnormal{ren} + \sqrt{1-\mathbf{O}}\odot \epsilon \,, \quad \quad \epsilon \sim \mathcal{N}(\mathbf{0}, \mathbf{1}) \textnormal{.}
%\mathbf{F} =  \mathbf{O} \odot \mathbf{F}^\textnormal{ren} + (1-\mathbf{O})\odot \epsilon \,, \quad \quad \epsilon \sim \mathcal{N}(0, \mathbf{1}) \textnormal{.}
\end{equation}
The decoder can generate plausible details to fill these empty areas, since it is trained using a GAN formulation (c.f. Sec.~\ref{sec:training}).

\subsection{Encoding Reference Views}
\label{sec:encoder}
The variational 3D Gaussian representation described in the previous section is obtained from two given reference views $\mathbf{I}_1$, $\mathbf{I}_2$ sampled from a video sequence. To this end, we adapt the encoder from pixelSplat~\cite{charatan23pixelsplat} to our setting of variational Gaussians by adding the capability of predicting means $\mathbf{h}_\sigma$ and variances $\mathbf{h}_\sigma$ of spherical harmonic coefficients for view-dependent features of each predicted 3D Gaussian (c.f. Fig~\ref{fig:method}, left).
The encoder consists of three parts: (1) a vision transformer~\cite{dino}, (2) an epipolar transformer~\cite{He2020epipolar}, and (3) a per-pixel sampling of 3D Gaussians from predicted depth distributions~\cite{charatan23pixelsplat}, which are shortly outlined in the following. For more details about the encoder architecture, we refer to the supplementals, Charatan et al.~\cite{charatan23pixelsplat}, and He et al.~\cite{He2020epipolar}.

\vspace{0.3cm}
\noindent\textbf{Vision transformer.}
%\paragraph{Vision transformer}
The vision transformer is applied to both tokenized images $\mathbf{I}_1$ and $\mathbf{I}_2$ to obtain pixel-aligned feature maps. Compared to pixelSplat~\cite{charatan23pixelsplat}, we omit the ResNet and ony use a pre-trained DINO ViT-B/8~\cite{dino}.
Each of the outputs is annotated with depth values from epipolar lines of the other view~\cite{He2020epipolar}.

\vspace{0.3cm}
\noindent\textbf{Epipolar transformer.} 
%\paragraph{Epipolar transformer}
Epipolar cross attention~\cite{He2020epipolar} is used to allow communication of features across corresponding pixels from both views. To this end, the attached depth values from corresponding epipolar lines are positionally encoded and concatenated to the individual feature maps. Then, keys, queries and values are computed for all locations before performing cross attention between each pixel and samples from its epipolar line in the other view.
The communication between two views allows the encoder to resolve the scale ambiguity~\cite{charatan23pixelsplat}.

\vspace{0.3cm}
\noindent\textbf{Gaussian sampling head.}
%\paragraph{Gaussian sampling head}
Last, each final feature map is used to predict a distribution over its rays, indicating the probability of a 3D Gaussian lying at the specific depth~\cite{charatan23pixelsplat}. To that end, each ray is divided into a set of bins over which a discrete distribution is predicted. Further, for each bin, an offset is predicted to allow for obtaining accurate positions in 3D space. Multiple Gaussians can be sampled per ray. The probability of a sampled Gaussian is used as opacity $o$. For each sample, we also predict the remaining Gaussian properties of scale $\mathbf{S}$, rotation $\mathbf{R}$ (as quaternion), color $\mathbf{c}$, and variational parameters $(\mathbf{h}_\mu, \mathbf{h}_\sigma)$.

\vspace{0.3cm}\noindent
With the described encoder, uncertainty is modeled in two ways: First, the Gaussian locations are sampled from the predicted distributions over the rays, modeling uncertainty in 3D Gaussian location. Second, uncertainty in local appearance is modeled via distribution parameter prediction of the variational Gaussians. 

\subsection{Decoding}
\label{sec:decoder}
We render both, RGB colors and features into pixel space using the adapted 3D Gaussian rasterizer (c.f. Sec.~\ref{sec:variational_gaussians}). We found that also formulating a loss directly on an RGB output helps with optimizing the structural Gaussian parameters position $\mathbf{x}$, scale $\mathbf{S}$, rotation $\mathbf{R}$, and opacity $o$. To interpret the features, we use the pre-trained light-weight VAE decoder from LDM~\cite{stable_diffusion} (c.f. Fig.~\ref{fig:method}, right). It is a purely convolutional architecture with four upsample blocks, each consisting of two residual blocks. The decoder receives multi-scale feature images: first, we bilinearly down-sample the rendered feature image three times along the spatial dimensions. Then, we feed the different scales into the U-Net decoder at different stages of the architecture. The decoder is trained together with the remaining architecture using reconstruction and generative losses, as described in Sec.~\ref{sec:training}.

\subsection{Training}
\label{sec:training}
The presented architecture consisting of encoder, variational Gaussians, and decoder is trained in an end-to-end fashion on video data. In each iteration, we sample a video from our training dataset, select two reference views and four target views. The selection criteria differ for different scene types (large scenes and object-centric scenes) and are detailed in the experimental setup in Sec.~\ref{sec:experimental_setup}. The reference views are encoded into variational 3D Gaussians, rendered from the target camera perspectives, and decoded using the VAE decoder. We train all networks using the following losses.

\vspace{0.3cm}
\noindent\textbf{Reconstruction losses.}
%\paragraph{Reconstruction Losses}
Similar to LDM~\cite{stable_diffusion}, we use a combination of a standard L1 loss and LPIPS as a perceptual loss between the decoded feature image $\hat{\mathbf{T}}$ and the target image $\mathbf{T}$: 
\begin{equation}
\mathcal{L}_\mathrm{rec} =  \lambda_1\|\hat{\mathbf{T}}-\mathbf{T}\|_{1} + \lambda_2~\mathrm{LPIPS}(\hat{\mathbf{T}},\mathbf{T})
\end{equation}

\vspace{0.3cm}
\noindent\textbf{Auxiliary losses.}
%\paragraph{Auxiliary Losses}
We further apply an auxiliary loss directly between the color renderings $\hat{\mathbf{T}}_{\mathrm{aux}}$ and the target image $\mathbf{T}$ to provide better gradients to the structural parameters (c.f. Sec~\ref{sec:variational_gaussians}):
\begin{equation}
\mathcal{L}_\mathrm{aux} = \lambda_3\|\hat{\mathbf{T}}_{\mathrm{aux}}-\mathbf{T}\|_{2}^2 + \lambda_4~\mathrm{LPIPS}(\hat{\mathbf{T}}_{\mathrm{aux}},\mathbf{T})
\end{equation}

\vspace{0.3cm}
\noindent\textbf{Generative loss.}
%\paragraph{Generative Loss}
To enable correct sampling and generation in uncertain regions, we further optimize our method using GAN losses, adapted from the LDM VAE-GAN decoder~\cite{stable_diffusion}. We add and further train a pre-trained discriminator network~\cite{Isola2017CVPR} $D$ from LDM\cite{stable_diffusion} predicting the likelihood of image patches being real. Therefore, our architecture is trained to directly maximize its output:
\begin{equation}
    \mathcal{L}_{\mathrm{gen}}=E_{\hat{\mathbf{T}}} [\log (1- D(\hat{\mathbf{T}}))]\textnormal{,}  \quad \quad \mathcal{L}_\textnormal{disc} = E_{\mathbf{X}}[\log(D(\mathbf{X}))] + E_{\hat{\mathbf{T}}}[\log(1-D(\hat{\mathbf{T}}))]\textnormal{,}
\end{equation}
where $\mathbf{X}$ are real images from the training dataset.
In summary, our autoencoder is optimized with $\mathcal{L}=\mathcal{L}_\mathrm{rec}+\mathcal{L}_\mathrm{aux}+\mathcal{L}_{\mathrm{gen}}$ and the discriminator with $\mathcal{L}_\textnormal{disc}$.

\section{Experiments}
\label{sec:experiments}
In this section, we detail our experiments made with latentSplat, provide comparisons with state-of-the-art baselines, and verify our architecture design in form of an ablation.
Our experiments aim to support the following statements:
(1) latentSplat improves on previous methods for two-view interpolation in terms of visual quality, (2) generalizes better to extrapolation, i.e., target views outside of reference views, (3) avoids unnecessary hallucination but sticks to the identity of the observed scene, (4) enables 3D reconstruction by predicting consistent novel views, and (5) maintains the real-time rendering and memory efficiency of 3D Gaussian splatting.
We provide additional results in the appendix.

\subsection{Experimental Setup}
\label{sec:experimental_setup}
\noindent\textbf{Datasets.}
%\paragraph{Datasets}
We conduct two-view reconstruction experiments for an object-centric setting and for general video datasets capturing diverse scenes.
For the former, we use the Common Objects in 3D~\cite{co3d} dataset, which consists of video captures of real-world objects grouped into categories. Following related work~\cite{chan2023genvs}, we choose cleaned subsets of hydrants and teddybears, randomly split into 95\% training and 5\% test data.
In all experiments, including baselines, we gradually increase the gap between reference views from initially 8-18 up to 25 frames roughly corresponding to 90\textdegree~in the 102 frame videos.
At the same time, we increasingly randomize the target view selection from pure interpolation until a uniform distribution over all views.
For evaluation on general scenes, we leverage RealEstate10k~\cite{Zhou2018multiplane}, a dataset of home walkthrough clips gathered from about 10000 YouTube videos. We use provided splits and adopt the training curriculum from pixelSplat~\cite{charatan23pixelsplat} except of increasing the sampling interval of target views up to 45 frames before and after both reference views to learn extrapolation.

\vspace{0.3cm}
\noindent\textbf{Baselines.}
%\paragraph{Baselines}
We compare our approach against four baselines.
pixelNeRF~\cite{yu2021pixelnerf} conditions a single NeRF MLP by interpolating pixel-aligned features of reference views. Du et al.~\cite{cross_attn} is a light field rendering approach that uses multi-view self-attention and cross-attention of target rays to samples along its epipolar lines in the input images.
pixelSplat\cite{charatan23pixelsplat} predicts 3D Gaussians along the rays of two reference views that can be efficiently rendered via rasterization.
Unlike the previous regression-based approaches, GeNVS~\cite{chan2023genvs} proposes a generative diffusion model with view-conditioning via pixelNeRF in a feature space.

\begin{table*}[!t]

\caption{\textbf{360° Novel view synthesis on CO3D.} We outperform previous methods in terms of generative metrics and perceptual metrics. Due to the generative nature of our method, we are only on-par in traditional reconstruction metrics of PSNR and SSIM, as they strongly punish generation of details if they do not match the target. 
}
\label{tab:quantitative_co3d}

\centering
\begingroup
\sisetup{
  table-align-uncertainty=true,
  separate-uncertainty=true,
}
%% local redefinitions
\renewrobustcmd{\bfseries}{\fontseries{b}\selectfont}

\resizebox{\textwidth}{!}{%
\begin{tabular}{c|l|cc|cc|cc|cc|cc|cc}
\toprule
&& \multicolumn{6}{c|}{Interpolation} & \multicolumn{6}{c}{Extrapolation} \\ 
Cat. & Method & FID$\downarrow$ & KID$\downarrow$ & LPIPS$\downarrow$ & DISTS$\downarrow$ & PSNR$\uparrow$ & SSIM$\uparrow$ & FID$\downarrow$ & KID$\downarrow$ & LPIPS$\downarrow$ & DISTS$\downarrow$ & PSNR$\uparrow$ & SSIM$\uparrow$ \\ 
\midrule
\midrule
& pixelNeRF~\cite{yu2021pixelnerf} & 183.24 & 0.104 & 0.566 & 0.345 & \underline{18.39} & 0.411 & 238.41 & 0.156 & 0.658 & 0.429 & \textbf{16.24} & \underline{0.360} \\ 
& Du et al.~\cite{cross_attn} & 154.20 & 0.090 & 0.471 & 0.276 & \textbf{18.78} & \textbf{0.476} & 275.61 & 0.208 & 0.599 & 0.382 & \underline{15.79} & \textbf{0.366} \\
& pixelSplat~\cite{charatan23pixelsplat} & \underline{58.13} & \underline{0.011} & \underline{0.401} & \underline{0.203} & 18.05 & \underline{0.429} & \underline{92.61} & \underline{0.030} & \underline{0.485} & \underline{0.263} & 15.75 & 0.332 \\ 

\rowcolor{rowhighlight}%
\cellcolor{white}\parbox[t]{2mm}{\multirow{-4}{*}{\rotatebox[origin=c]{90}{Hydrants}}} & \textbf{Ours} & \textbf{40.93} & \textbf{0.005} & \textbf{0.356} & \textbf{0.166} & 18.01 & 0.413 & \textbf{48.03} & \textbf{0.008} & \textbf{0.426} & \textbf{0.202} & 15.78 & 0.306 \\

\midrule
& pixelNeRF~\cite{yu2021pixelnerf} & 179.85 & 0.082 & 0.580 & 0.386 & 18.97 & 0.580 & 236.82 & 0.132 & 0.649 & 0.450 & 17.05 & 0.531 \\ 
& Du et al.~\cite{cross_attn} & 141.16 & 0.065 & 0.436 & 0.245 & 20.69 & \underline{0.666} & 229.78 & 0.142 & 0.564 & 0.347 & 16.65 & \underline{0.553} \\
& pixelSplat~\cite{charatan23pixelsplat} & \underline{74.82} & \underline{0.014} & \underline{0.369} & \underline{0.200} & \underline{20.73} & \textbf{0.687} & \underline{123.33} & \underline{0.047} & \underline{0.473} & \underline{0.260} & \underline{17.51} & \textbf{0.564} \\ 

\rowcolor{rowhighlight}%
\cellcolor{white}\parbox[t]{2mm}{\multirow{-4}{*}{\rotatebox[origin=c]{90}{Teddybears}}} & \textbf{Ours} & \textbf{53.47} & \textbf{0.004} & \textbf{0.338} & \textbf{0.173} & \textbf{20.83} & 0.663 & \textbf{71.12} & \textbf{0.010} & \textbf{0.434} & \textbf{0.219} & \textbf{17.71} & 0.533 \\
\bottomrule

\end{tabular}

}
\endgroup
\end{table*}

\begin{figure}[!t]
    \centering
    \includesvg[width = 1.06\textwidth]{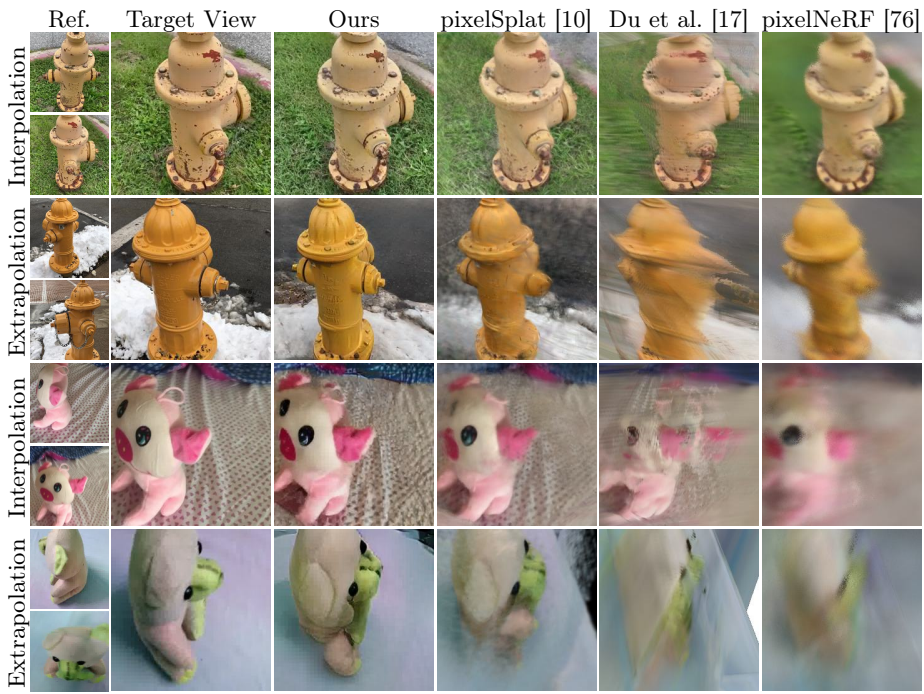}
    \caption{\textbf{Qualitative results on the CO3D dataset~\cite{co3d}.} We evaluate two-view NVS on hydrants and teddybears. latentSplat synthesizes high-quality 360\textdegree~novel views, whereas regression-based approaches suffer from uncertainty resulting in blur.}
    \label{fig:qualitative_co3d}
    %\vspace{-0.4cm}
\end{figure}

\vspace{0.3cm}
\noindent\textbf{Metrics.}
%\paragraph{Metrics}
We employ three groups of each two metrics:
(1) FID~\cite{Heusel2017fid} and KID~\cite{binkowski2018demystifying} measure the similarity between the distributions of predicted novel views and the corresponding ground truth. They are the established metrics for image synthesis of generative models and reflect visual quality.
(2) Perceptual metrics like LPIPS~\cite{lpips} and DISTS~\cite{dists} leverage features of deep networks for the comparison of images w.r.t. structure and texture.
(3) Classical reconstruction metrics such as PSNR and SSIM are still well-established for evaluation of dense-view 3D reconstruction. However, because incorporating plain pixel-wise similarities, these metrics prefer blur over realistic details and are therefore not well-suited for the evaluation of generative methods.

\noindent
Further details can be found in the supplementary materials and our code.

\subsection{Object-Centric 3D Reconstruction}
\label{sec:object_centric}
Table~\ref{tab:quantitative_co3d} summarizes our quantitative results for object-centric 3D reconstruction on CO3D\cite{co3d}.
We outperform all baselines by a large margin in FID and KID while also significantly improving upon pixelSplat~\cite{charatan23pixelsplat} in perceptual metrics.
This indicates that we achieve a better visual quality and at the same time remain faithful w.r.t. to the observed scene.
Despite being a generative approach, we also outperform all baselines in PSNR on teddybears, but fall short in SSIM.
However, our deterministic ablation (c.f. Sec.~\ref{sec:ablations}) validates that these classical metrics prefer blurry reconstructions over perceptually good ones with generated details.
The differentiation of interpolation and extrapolation reveals that our approach generalizes better to unseen areas of the scene, as shown by a smaller performance difference compared to the baselines.
We show qualitative results in Fig.~\ref{fig:qualitative_co3d}.
latentSplat predicts sharp and detailed reconstructions fitting to the observations.
Moreover, it succeeds in synthesizing completely unobserved areas allowing full 360° generalization.
For the challenging case of modeling the unobserved backside of a hydrant (c.f. third row in Fig.~\ref{fig:qualitative_co3d}), all regression-based approaches predict a blurry reconstruction indicating high uncertainty. latentSplat's ability of generating a realistic novel view in this case highlights the advantage of uncertainty and generation in the rendering process.

\begin{table*}[t]
% \small
\caption{\textbf{Novel view synthesis on RE10k.} We compare against recent methods on view interpolation and extrapolation on the large scale RealEstate10k dataset~\cite{Zhou2018multiplane}, showing that our method can handle large scale scenes and high resolution images. Same as in the CO3D setting, we outperform previous works in generative metrics and perceptual metrics, while being on par in traditional metrics. Qualitative results in Fig.~\ref{fig:qualitative_re10k} show that we produce higher quality reconstructions, especially for extrapolation.}
\label{tab:quantitative_re10k}

\centering

\begingroup
\sisetup{
  table-align-uncertainty=true,
  separate-uncertainty=true,
}
%% local redefinitions
\renewrobustcmd{\bfseries}{\fontseries{b}\selectfont}

\resizebox{\textwidth}{!}{%
\begin{tabular}{l|cc|cc|cc|cc|cc|cc}
\toprule
& \multicolumn{6}{c|}{Interpolation} & \multicolumn{6}{c}{Extrapolation} \\ 
Method & FID$\downarrow$ & KID$\downarrow$  & LPIPS$\downarrow$ & DISTS$\downarrow$ & PSNR$\uparrow$ & SSIM$\uparrow$ & FID$\downarrow$ & KID$\downarrow$ & LPIPS$\downarrow$ & DISTS$\downarrow$ & PSNR$\uparrow$ & SSIM$\uparrow$ \\ 
\midrule
\midrule
pixelNeRF~\cite{yu2021pixelnerf} & 152.16 & 0.132 & 0.550 & 0.359 & 20.51 & 0.592 & 160.77 & 0.141 & 0.567 & 0.371 & 20.05 & 0.575 \\ 
Du et al.~\cite{cross_attn} & 9.73 & 0.005 & 0.219 & 0.133 & \textbf{24.55} & \underline{0.812} & 11.34 & 0.006 & 0.242 & 0.144 & 21.83 & \textbf{0.790} \\
pixelSplat~\cite{charatan23pixelsplat} & \underline{4.41} & \underline{0.002} & \underline{0.174} & \underline{0.107} & \underline{24.32} & \textbf{0.822} & \underline{5.78} & \underline{0.003} & \underline{0.216} & \underline{0.130} & \underline{21.84} & \underline{0.777} \\ 
\rowcolor{rowhighlight}%
\textbf{Ours} & \textbf{2.22} & \textbf{0.001} & \textbf{0.164} & \textbf{0.094} & 23.93 & \underline{0.812} & \textbf{2.79} & \textbf{0.001} & \textbf{0.196} & \textbf{0.109} & \textbf{22.62} & \underline{0.777} \\
\bottomrule
\end{tabular}
}
\endgroup
\label{tab:quantitative_re10k}
%\vspace{-0.3cm}
\end{table*}

\subsection{Scene-Level 3D Reconstruction}
\label{sec:scene_level}
\begin{figure}[!t]
    \centering
    \includesvg[width = 1.06\textwidth]{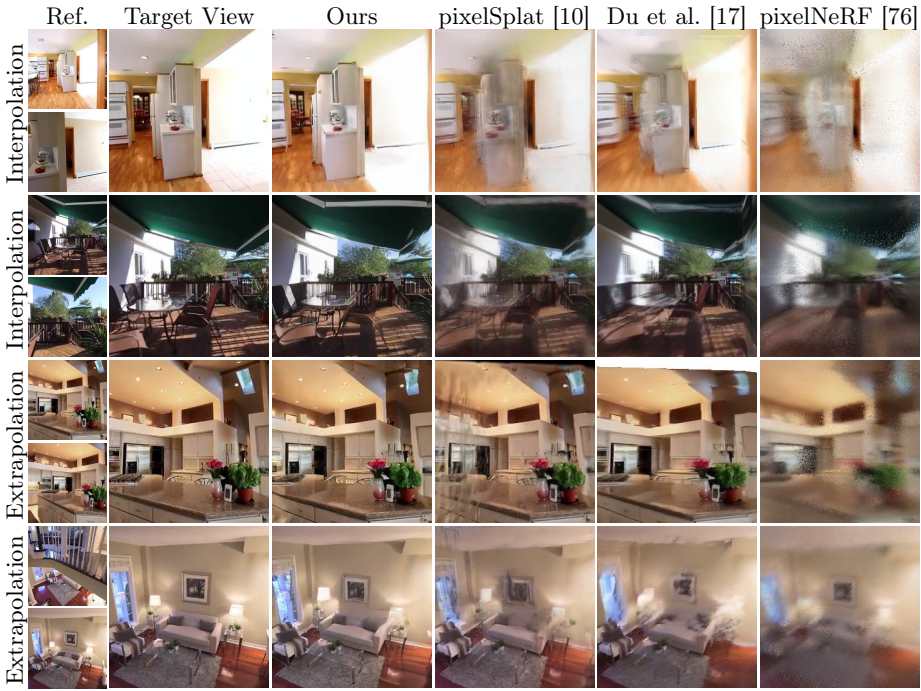}
    \caption{\textbf{Qualitative results on RealEstate10k}~\cite{Zhou2018multiplane}. We show that in many challenging cases latentSplat outperforms previous methods. This specifically holds for (1) reference views far apart from each other or (2) extrapolation outside of reference views.}
    \label{fig:qualitative_re10k}
    %\vspace{-0.2cm}
\end{figure}

We report quantitative results for scene-level 3D reconstruction on RealEstate10k in Table~\ref{tab:quantitative_re10k}.
Again, latentSplat outperforms all baselines in FID and KID as well as LPIPS and DISTS.
Although extrapolation on scene level is a quite different task compared to learning a category-level prior for object-centric videos, we observe the same result that latentSplat generalizes better to extrapolation compared to all baselines, even achieving state-of-the-art in PSNR as well.
This highlights the applicability of our method for various kinds of real-world videos.
Looking at qualitative examples in Fig.~\ref{fig:qualitative_re10k}, we can see that our approach produces clean and visually pleasing novel views with significantly less artifacts.

\subsection{Uncertainty Visualization}
\label{sec:uncertainty}
We aim to illustrate the uncertainty of our variational Gaussians directly by rendering the standard deviation $\mathbf{h}_\sigma$ in Eq.~\ref{eq:sampling}, averaged over all feature channels. To deal with empty regions outside of the reference camera frustums, we set the background standard deviation to one, which is in line with our feature map sampling in Eq.~\ref{eq:rendering}.
The resulting images (second row of Fig.~\ref{fig:intermediate}) show generally higher uncertainty (dark) for the background, which is either completely invisible or only partly visible in the reference views, compared to the main object, for which the model learns a category-level prior.
For the foreground, the model is less certain about details like edges or the fur of teddybears than about plain uniform surfaces, which explains the advantage of the generative decoder w.r.t. a higher level of detail. We provide more examples in the supplementary materials.

\subsection{Mesh Reconstruction from Novel Views}
\label{sec:mesh_reconstruction}

\begin{figure}[!t]
  \centering
    \begin{subfigure}[t]{0.32\textwidth}
    \includegraphics[width=1\linewidth]{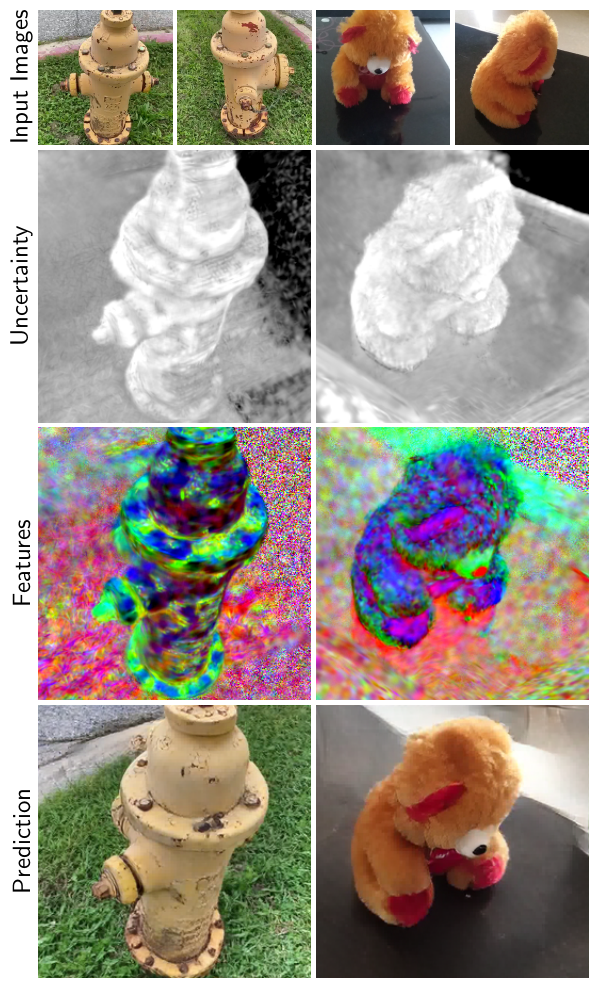}
    \caption{\textbf{Intermediate results.}}
    \label{fig:intermediate}
  \end{subfigure}
  \hfill
  \begin{subfigure}[t]{0.67\textwidth}
    \includegraphics[width=1\linewidth]{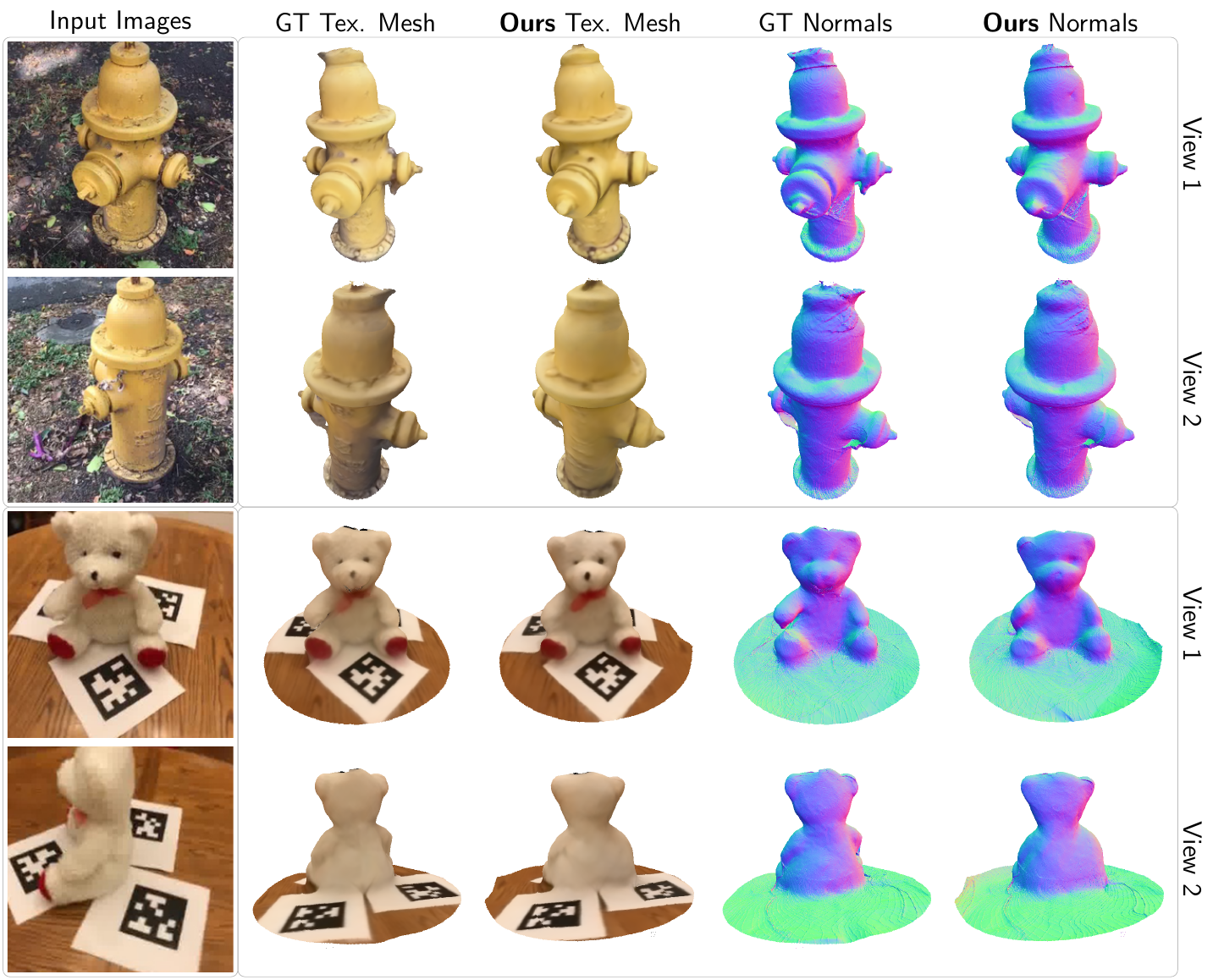}
    \caption{\textbf{Mesh reconstruction.}}
    \label{fig:mesh_reconstruction}
  \end{subfigure}
  \caption{\textbf{a)} Top to bottom: Given two \textit{input images}, latentSplat predicts variational Gaussians that model local \textit{uncertainty} in 3D. Rendering uncertainty reveals a focus on foreground details and out-of-view background. Sampled semantic Gaussians are rendered to \textit{features} and decoded to final \textit{predictions}. \textbf{b)} We apply mesh reconstruction on dense synthesized novel views. latentSplat is able to closely approximate the texture and geometry that we can obtain from reconstruction with ground-truth 360° videos.
  }
\end{figure}

High-quality novel view synthesis does not imply strong 3D reconstruction. Especially, considering generative methods, realistic novel views may not be 3D consistent.
Therefore, we examine latentSplat's abilities for downstream 3D reconstruction.
Starting with two input images, we sample semantic Gaussians once and render them from all camera poses of the original video.
Given these synthesized views, we can employ any surface reconstruction method~\cite{neus, volsdf, 2dgs} to obtain textured meshes. 
We use 2D Gaussian Splatting~\cite{2dgs}. 
The same procedure is applied for the original video and pixelSplat~\cite{charatan23pixelsplat} outputs to obtain ground-truth and baseline meshes, respectively.
We provide quantitative results for Chamfer distance in Table~\ref{tab:mesh_co3d}.
Our outputs are better suited for reconstruction. This indicates that probabilistic modeling is helpful for faithful surface reconstruction.
Qualitative results in Fig.~\ref{fig:mesh_reconstruction} further demonstrate that reconstructions with latentSplat are close to the ground truth w.r.t. texture and surface normals.

%\vspace{-0.2cm}

\begin{table}[t!]
\caption{\textbf{3D reconstruction on CO3D.} Chamfer distance ($\times 10^{-8}$): Gaussians / mesh from output images versus mesh from ground-truth videos.}

\centering
\begingroup
\sisetup{
  table-align-uncertainty=true,
  separate-uncertainty=true,
}
%% local redefinitions
\renewrobustcmd{\bfseries}{\fontseries{b}\selectfont}
\scriptsize
%\underline and %\textbf
%\textemdash for empty ones
\begin{tabular}{l|c|c}
\toprule
Method & Hydrants & Teddybears \\ 
\midrule
pixelSplat (Gaussians) & 49.343 & 24.965 \\
pixelSplat (mesh from images) & 1.815 & 1.905 \\
\rowcolor{rowhighlight}%
\textbf{Ours} (mesh from images)& \textbf{1.535} & \textbf{1.504} \\
\bottomrule
\end{tabular}
\endgroup
\label{tab:mesh_co3d}
\end{table}

\subsection{Efficiency}
\label{sec:runtime}
Table~\ref{tab:efficiency} shows our time and memory requirements for training and inference compared to the baselines.
Compared to pixelSplat~\cite{charatan23pixelsplat}, our encoding is slightly faster while the rendering is only a 1ms slower stemming to 68\% from the convolutional decoder and 32\% from splatting additional feature channels. Hence, we maintain the real-time rendering of 3D Gaussian splatting despite introducing a generative model. Furthermore, we are memory efficient during both training and inference. Compared to GeNVS~\cite{chan2023genvs}, we are orders of magnitude faster.

\subsection{Ablations}
\label{sec:ablations}
We conduct an ablation study for view extrapolation on CO3D hydrants with the results given in Table~\ref{tab:ablation}.
Interestingly, while performing much worse in FID and KID with a deterministic version omitting the variational formulation as well as the GAN loss, we obtain state-of-the-art results for PSNR and SSIM. This shows the trade-off between conservative and therefore blurry reconstructions favored by classical metrics and realistic and detailed novel views rewarded by FID and KID.
%Without latent and/or RGB skip connections, we perform slightly worse in most metrics validating our design choice of leveraging the rendered features and colors at multiple resolutions.
Overall, we find the best balance in metric scores by using our generative approach together with multi-resolution latent and RGB skip connections. We attribute that to the decoder having global and local context when interpreting features for image synthesis, which helps generating consistent results.

\begin{table*}[t]

\caption{\textbf{Ablation study.} 
We ablate our architecture design for extrapolation on CO3D hydrants. \emph{Deterministic} shows a version of our architecture without the variational formulation and without the GAN loss. \emph{No skip con.} omits the multi-scale decoder input and \emph{No RGB skip} omits the rendered RGB input to the decoder.
}
\label{tab:ablation}
\centering
\begingroup
\sisetup{
  table-align-uncertainty=true,
  separate-uncertainty=true,
}
%% local redefinitions
\renewrobustcmd{\bfseries}{\fontseries{b}\selectfont}
\scriptsize
%\underline and %\textbf
\begin{tabular}{l|cc|cc|cc}
\toprule
Method & FID$\downarrow$ & KID$\downarrow$  & LPIPS$\downarrow$ & DISTS$\downarrow$ & PSNR$\uparrow$ & SSIM$\uparrow$ \\ 
\midrule
\midrule
Deterministic & 69.33 & 0.022 & \textbf{0.406} & 0.240 & \textbf{16.50} & \textbf{0.340} \\
No skip con. & 50.31 & \underline{0.007} & 0.440 & \textbf{0.204} & 15.55 & 0.293 \\
No RGB skip & \textbf{48.48} & \textbf{0.006} & 0.438 & 0.206 & 15.48 & 0.288 \\
%No VAE & \textbf{48.45} & \textbf{0.006} & 0.437 & 0.207 & 15.59 & 0.290 \\
\rowcolor{rowhighlight}%
\textbf{Ours} & \underline{48.88} & \textbf{0.006} & \underline{0.436} & \underline{0.205} & \underline{15.61} & \underline{0.299} \\
\bottomrule
\end{tabular}
\endgroup
%\vspace{-0.3cm}
\end{table*}

\begin{table*}[t]

\caption{\textbf{Efficiency comparison.} Our method renders roughly 2000 times faster and is much less expensive to train than the state-of-the-art generative model GeNVS~\cite{chan2023genvs}. Compared to the regression-based pixelSplat~\cite{charatan23pixelsplat}, latentSplat entails a negligible increase in resource requirements in turn for significantly better image quality.}
\label{tab:efficiency}

\centering
\begingroup
\sisetup{
  table-align-uncertainty=true,
  separate-uncertainty=true,
}
%% local redefinitions
\renewrobustcmd{\bfseries}{\fontseries{b}\selectfont}
\scriptsize
%\underline and %\textbf
\begin{tabular}{l|cc|c|c}
\toprule
& \multicolumn{2}{c|}{Inference Time (s)} & Training Time & Inference \\ 
Method & Encode$\downarrow$ & Render$\downarrow$ & (GPU-h)$\downarrow$ & Memory (GB)$\downarrow$ \\ 
\midrule
\midrule
pixelNeRF~\cite{yu2021pixelnerf} & \textbf{0.003} & 5.464 & \textbf{96} & 3.961 \\
Du et al.~\cite{cross_attn} & 0.011 & 1.337 & 288 & 19.604 \\
pixelSplat~\cite{charatan23pixelsplat} & 0.113 & \textbf{0.002} & \underline{192} & \textbf{2.755} \\
GeNVS~\cite{chan2023genvs} & - & 6.423\footnotemark[1] & 2112 & - \\
\rowcolor{rowhighlight}%
\textbf{Ours} & \underline{0.080} & \underline{0.003} & \underline{192} & \underline{3.161} \\
\bottomrule
\end{tabular}
%}
\endgroup
\end{table*}

\footnotetext[1]{GeNVS~\cite{chan2023genvs} does not provide code. We estimated the rendering time based on pixelNeRF\cite{yu2021pixelnerf} and StableDiffusion~\cite{stable_diffusion} denoising for 25 steps and resolution 256.}

\section{Conclusion}
We presented latentSplat, a method that successfully combines the strengths of regression-based approaches with the power of a lightweight generative model to handle uncertainty. Our approach achieves state-of-the-art quality in novel view synthesis from two input images while providing the highest perceptual similarity to the ground truth. Predicted views are 3D consistent, enabling downstream mesh reconstruction. Compared to previous generative approaches, latentSplat is much faster and more scalable, allowing real-time rendering in large resolutions.

\vspace{0.3cm}
\noindent\textbf{Limitations and future work.} 
%\paragraph{Limitations and Future Work}
Since the location of Gaussians is limited to the camera frustums of input views, the 2D decoder inpaints out-of-view areas for extrapolation resulting in 3D inconsistencies. Another limitation is the \textit{independent} sampling of depth along rays and local appearance from variational Gaussians. Finding a trade-off between capturing the full conditional distribution of reconstructions and sampling efficiency renders potential future work.

\vspace{0.3cm}
\noindent\textbf{Acknowledgements.}
%\paragraph{Acknowledgements}
This project was partially funded by the \emph{Saarland/Intel Joint Program on the Future of Graphics and Media.} We also thank David Charatan and co-authors for the great pixelSplat codebase.

% \newpage
\bibliographystyle{splncs04}
\bibliography{arxiv}

\newpage

\title{latentSplat: Autoencoding Variational Gaussians \\ for Fast Generalizable 3D Reconstruction}
\subtitle{Supplementary Material}
\titlerunning{latentSplat: Variational Gaussians}
\author{}
\authorrunning{C.~Wewer et al.}
\institute{}

\maketitle

\setcounter{section}{0}
\renewcommand\thesection{\Alph{section}}

%\section*{Content}
In this supplementary material, we give details regarding architecture and evaluation in Sections~\ref{sec:architecture_details} and~\ref{sec:evaluation_details}. We provide a separate comparison with the closest generative baseline in Sec.~\ref{sec:genvs}.
Section~\ref{sec:3d_consistency} deals with the evaluation of 3D consistency using videos.
We present visualizations of intermediate results as well as model uncertainty underlining the advantages of latentSplat as a fast generative approach in Section~\ref{sec:additional_visualizations}.
Finally, we provide additional qualitative results in Section~\ref{sec:additional_results}. Please also refer to the project website\footnote{\href{https://geometric-rl.mpi-inf.mpg.de/latentsplat/}{geometric-rl.mpi-inf.mpg.de/latentsplat/}} for results in motion.

%\begin{itemize}
%    \item implementation details
%        \begin{itemize}
%            \item architecture hyperparameters (latent dim, number of Gaussians per ray, degree of spherical harmonics, hidden dimensionalities of VAE...)
%            \item loss weights
%        \end{itemize}
%    \item training details
%        \begin{itemize}
%            \item training for how many iterations, days and how much VRAM?
%            \item benchmarking done on RTX 4090
%        \end{itemize}
%    \item evaluation details
%        \begin{itemize}
%            \item choice of evaluation examples (context and target views)
%        \end{itemize}
%    \item 3D consistency via videos and 360° NVS
%        \begin{itemize}
%            \item direct prediction given two views vs
%            \item autoregressive approach
%            \item 360° novel view synthesis on CO3D hydrants and teddybears if possible
%        \end{itemize}
%    \item uncertainty visualization if possible
%    \item more qualitative examples of everything
%\end{itemize}

\section{Architecture Details}
\label{sec:architecture_details}
Our architecture is composed of an encoder, a decoder, and a discriminator.

\vspace{0.3cm}
\noindent\textbf{Encoder} 
We adapt the encoder from pixelSplat~\cite{charatan23pixelsplat} to our setting. It consists of a pre-trained DINOv1 ViT-B/8~\cite{dino} backbone, an epipolar transformer~\cite{He2020epipolar}, and a Gaussian sampling head.
We upscale the local patch tokens from the backbone by repeating them for the original patch size and add the linearly projected global class token to obtain a feature map in the orginal image resolution.
The epipolar transformer has two blocks with four-head attention~\cite{He2020epipolar} to $32$ samples of each pixel's epipolar line. The token dimensionality is $128$ and for computational feasibility we downscale the input feature maps by a factor of $4$ in each dimension.
To prevent a bottleneck with low spatial dimensions in form of the epipolar transformer, we add a skip connection in the original resolution.
Finally, the Gaussian sampling head employs simple linear projections with activation functions specific to the individual parameters of the Gaussians like normalized quaternions for rotation and scaled and shifted sigmoid for scale.
The resulting variational Gaussians exhibit spherical harmonic coefficients up to a degree of $4$ for RGB and a Gaussian distribution of spherical harmonic coefficients up to a degree of $2$ for features of dimensionality $4$ (LDM VAE~\cite{stable_diffusion} feature size).

\vspace{0.3cm}
\noindent\textbf{Decoder}
We adapt the pre-trained VAE decoder from LDM~\cite{stable_diffusion}.
It is a fully convolutional architecture consisting of four upsample blocks, each consisting of two residual blocks. We add 1x1 convolutions as skip connections to feed in downscaled versions of the intermediate auxiliary RGB renderings and the feature maps to each upsample block. In order to avoid noisy gradients, we initialize the weights of these additional layers to zero. 

\vspace{0.3cm}
\noindent\textbf{Discriminator}
We finetune the pre-trained discriminator from LDM~\cite{stable_diffusion}.
It is a very small CNN with three strided convolutions as hidden layers such that it does not have a global receptive field and only predicts the likelihood of patches~\cite{Isola2017CVPR}.

\vspace{0.3cm}
\noindent\textbf{Loss}
Our loss can be split into auxiliary losses, reconstruction losses, and a generative loss.
We start the training with auxiliary losses only for the first 100k iterations to avoid fitting features to Gaussians in wrong positions.
Furthermore, we delay the generative loss until 125k iterations to prevent an overly strong discriminator.
To account for different scale, the individual terms have different weights.
As reconstruction losses, the weights of L1 and LPIPS are $\lambda_1=1$ and $\lambda_2=1$, respectively.
The mean squared error and LPIPS for the axiliary rendering are weighted by $\lambda_3=10$ and $\lambda_4=0.5$.
Finally, we adopt the adaptive weighting of the generative loss from LDM~\cite{stable_diffusion} by leveraging the ratio of gradient norms from reconstruction and generative losses to the last layer of the decoder.

\section{Evaluation Details}
\label{sec:evaluation_details}
\noindent\textbf{View Selection.}
For the evaluation on a test split, we create examples out of two reference and three target views.
For each scene of CO3D~\cite{co3d}, we sample $20$ reference view pairs within a frame distance of $16$ to $25$ while respecting the circular camera motion. Then we sample random target views within or outside of the reference views for interpolation or extrapolation, respectively.
For RealEstate10k~\cite{Zhou2018multiplane}, we stick to the same evaluation setup of pixelSplat~\cite{charatan23pixelsplat}. For each scene, we sample one reference view pair within a frame distance of $45$ to $135$ that fulfills the requirement of mutual overlap (coverage of epipolar lines) of at least 60\%.
For extrapolation, we sample target views up to $45$ frames previous/after the first/second reference view.

\vspace{0.3cm}
\noindent\textbf{Efficiency Comparison.}
We train all models including baselines until convergence, which translates to 200k iterations for latentSplat, pixelSplat~\cite{charatan23pixelsplat}, and pixelNeRF~\cite{yu2021pixelnerf}, and 300k for the method of Du et al.~\cite{cross_attn}. In wallclock time, this means we used two NVidia A40 GPUs for four days for both latentSplat and pixelSplat, one GPU for four days for pixelNeRF, and four GPUs for three days for the method of Du et al.~\cite{cross_attn}.
For a fair comparison in terms of inference time and memory, we evaluate all methods on the same data (extrapolation on CO3D hydrants) and the same device (NVidia RTX 4090).

\section{Comparison with GeNVS}
\label{sec:genvs}
\begin{figure}[t]
    \centering
    \includesvg[width=1.06\textwidth]{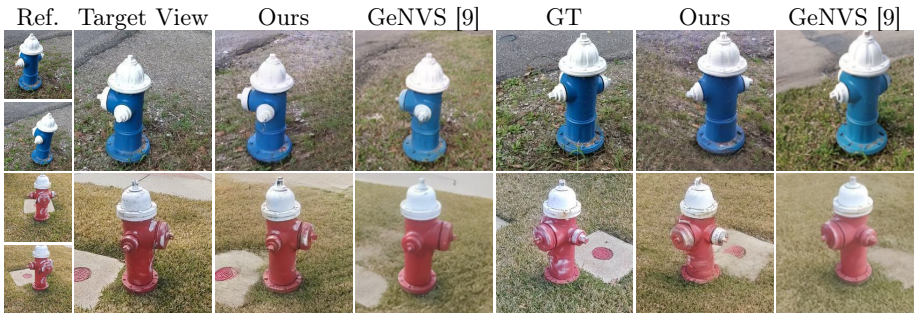}
    \caption{\textbf{Comparison with GeNVS~\cite{chan2023genvs}} We qualitatively compare against GeNVS. Note that the setups of both methods differ strongly and that GeNVS code is not available for reproducing results. Thus, this comparison is \textbf{not fair}. We selected the same CO3D test examples and compared against renderings shown in their paper. Both methods generate similar quality while ours is much faster (c.f. Sec. 4.6 main paper).}
    \label{fig:genvs}
    %\vspace{-0.3cm}
\end{figure}

We provide an additional qualitative comparison with GeNVS~\cite{chan2023genvs} in Fig.~\ref{fig:genvs}. Their code is not available and we are not able to reproduce their results for a fair comparison. We still aim to do a comparison using figures from their paper. Note that there are major differences between the setups of GeNVS and latentSplat such as the number of input views and the version of the dataset (CO3Dv1 vs CO3Dv2). We match their setting as close as possible by creating a test split of all their evaluation examples, finding closest poses, and using the same input view together with a close second view. Although the comparison of single- and two-view 3D reconstruction is not fair, note that GeNVS uses scale normalization and depth cues (c.f. supplementary material C.3) to resolve the scale ambiguity from a single input view.
The qualitative comparison shows similar image quality. However, our approach is much faster and efficient during both inference and training, as detailed in Sec. 4.6 of the main paper.

\section{3D Consistency}
\label{sec:3d_consistency}
We aim to assess the 3D consistency of novel view snythesis with latentSplat via a qualitative evaluation of videos. For this, please refer to the \href{https://geometric-rl.mpi-inf.mpg.de/latentsplat/}{project website}.
Given just two reference views, we are able to synthesize full 360° novel views for CO3D~\cite{co3d} hydrants and teddybears without obvious geometric inconsistencies.
Furthermore, the videos on RealEstate10k~\cite{Zhou2018multiplane} appear realistic without flickering from pixel-level differences in generation between nearby frames.
Note that we sample Gaussian features only once independent of the target views resulting in consistent renderings even in case of uncertainty.
Only low-opacity regions outside of the reference view frustums are filled with independently sampled noisy features in the target image space.
Hence, these especially difficult areas (usually background) are more prone to be inconsistent in different views.

\section{Additional Visualizations}
\label{sec:additional_visualizations}
\begin{figure}[!t]
    \centering
    \includesvg[width = \textwidth]{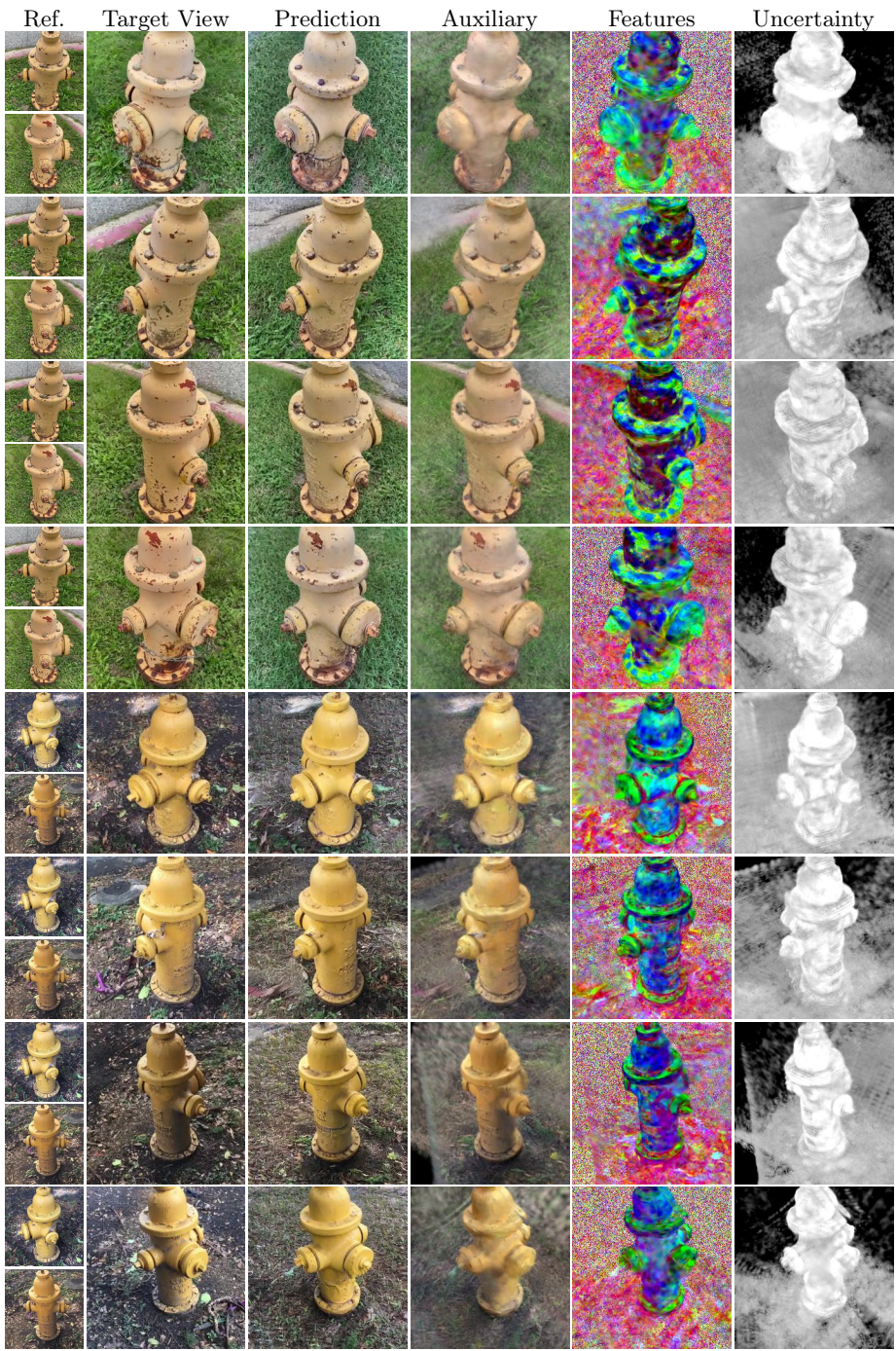}
    \caption{\textbf{Intermediate results for 360° novel view synthesis on CO3D hydrants~\cite{co3d}.} For uncertainty on the right, darker regions correspond to higher uncertainty. Features are visualized with PCA dimensionality reduction to 3 dimensions.}
    \label{fig:intermediate_co3d_hydrant_360}
\end{figure}

We visualize the process of assembling the final prediction in terms of intermediate results in Fig.~\ref{fig:intermediate_co3d_hydrant_360} for CO3D hydrants and Fig.~\ref{fig:intermediate_co3d_teddybear_360} for CO3D teddybears.

\vspace{0.3cm}
\noindent\textbf{Auxiliary.}
In order to provide better gradients to the structural parameters of the variational Gaussians, we apply an auxiliary reconstruction loss on directly rasterized RGB images skipping the VAE-GAN decoder.
These renderings (4th column) suffer from blur in regions of high uncertainty as well as artifacts like floating Gaussians.

\vspace{0.3cm}
\noindent\textbf{Features.}
Besides RGB, we render general feature maps that we visualize via employing PCA jointly over all pixel-aligned feature vectors of all frames (5th column). We observe that different parts of the objects are encoded by different latent features, while the background is clearly separated and filled with noise in areas of low density according to Eq.~3 in the main paper.

\vspace{0.3cm}
\noindent\textbf{Uncertainty.}
Finally, we aim to illustrate the uncertainty of our variational Gaussians directly by rendering the standard deviation $\mathbf{h}_\sigma$ in Eq.~2 of the main paper, averaged over all feature channels. To deal with empty regions outside of the reference camera frustums, we set the background standard deviation to one, which is in line with our feature map sampling in Eq.~3.
The resulting images (6th column) show generally higher uncertainty (dark) for the background, which is either completely invisible or only partly visible in the reference views, compared to the main object, for which the model learns a category-level prior.
For the main object, the model is less certain about details like edges or the fur of teddybears than about plain uniform surfaces, which explains the advantage of the generative decoder w.r.t. a higher level of detail.

\begin{figure}[!t]
    \centering
    \includesvg[width = \textwidth]{co3d_teddybear_360_features.svg}
    \caption{\textbf{Intermediate results for 360° novel view synthesis on CO3D teddybears~\cite{co3d}.} For uncertainty on the right, darker regions correspond to higher uncertainty. Features are visualized with PCA dimensionality reduction to 3 dimensions.}
    \label{fig:intermediate_co3d_teddybear_360}
\end{figure}

\section{Additional Results}
\label{sec:additional_results}
We provide additional qualitative results for interpolation and extrapolation on CO3D~\cite{co3d} and RealEstate10k~\cite{Zhou2018multiplane} in Fig.~\ref{fig:qualitative_co3d_hydrant_intra} to~\ref{fig:qualitative_re10k_extra}, and in the supplemental video.

\begin{figure}[!t]
    \centering
    \includesvg[width = 1.06\textwidth]{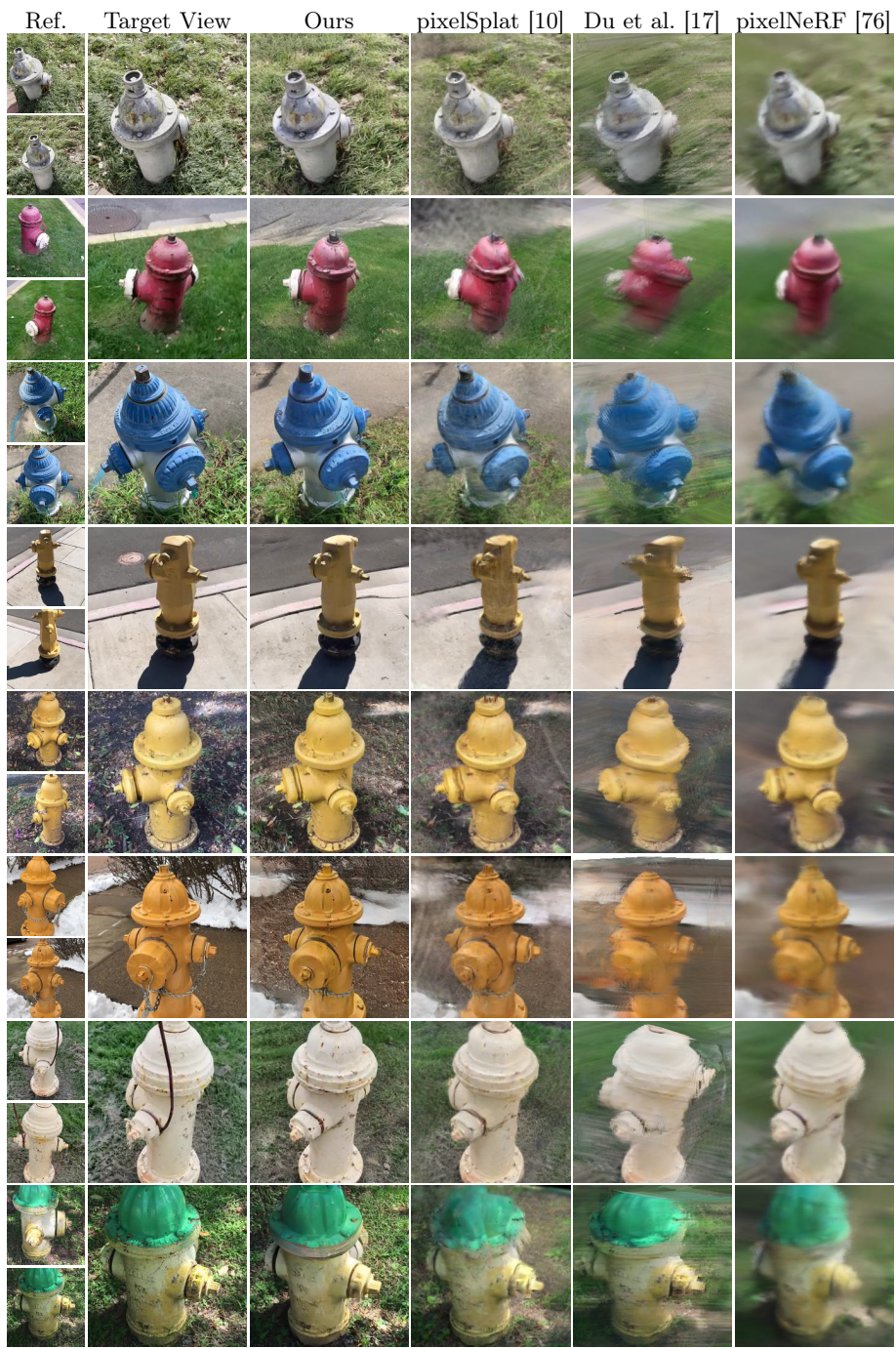}
    \caption{\textbf{Qualitative results for interpolation on CO3D hydrants~\cite{co3d}.}}
    \label{fig:qualitative_co3d_hydrant_intra}
\end{figure}

\begin{figure}[!t]
    \centering
    \includesvg[width = 1.06\textwidth]{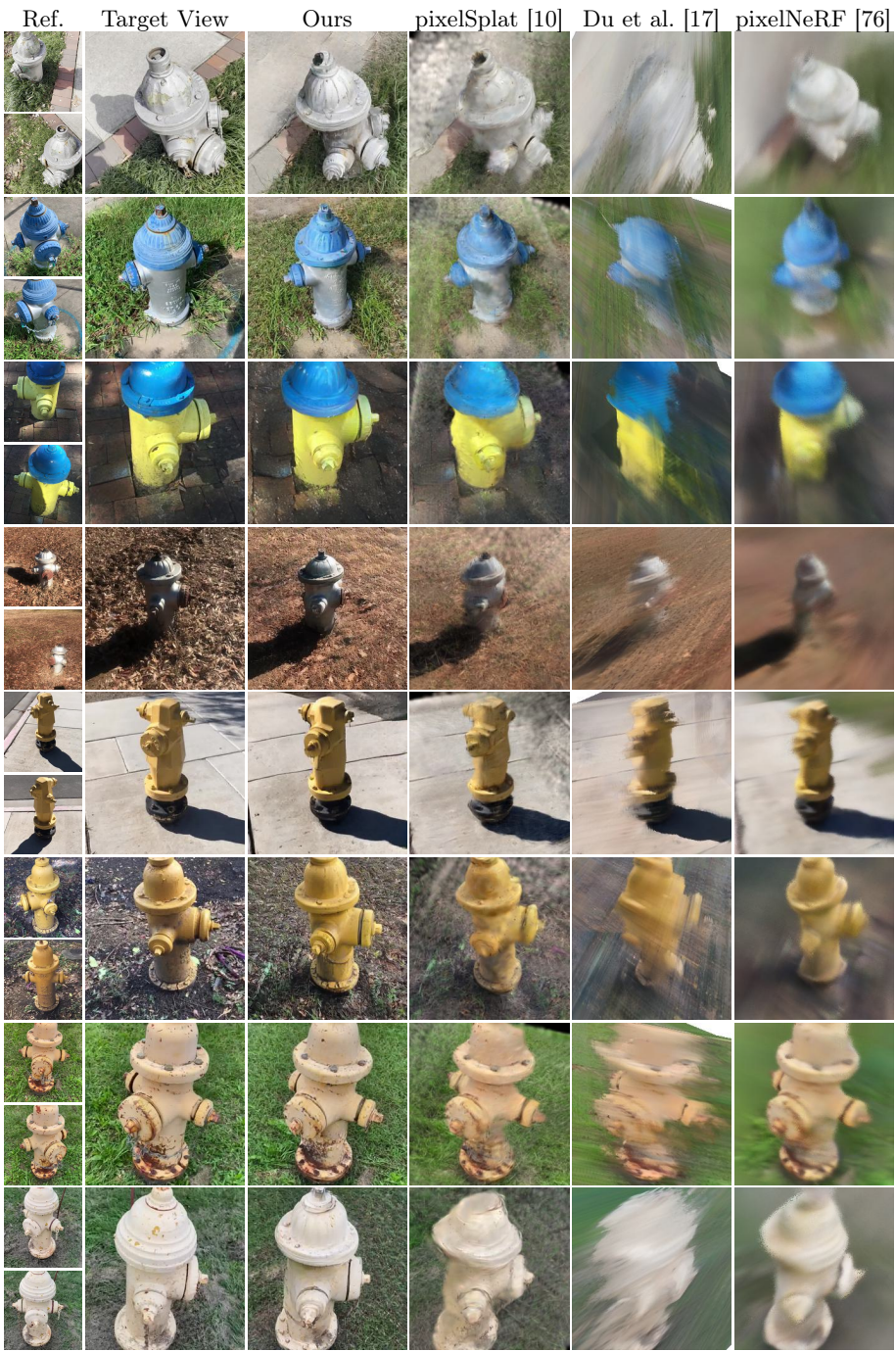}
    \caption{\textbf{Qualitative results for extrapolation on CO3D hydrants~\cite{co3d}.}}
    \label{fig:qualitative_co3d_hydrant_extra}
\end{figure}

\begin{figure}[!t]
    \centering
    \includesvg[width = 1.06\textwidth]{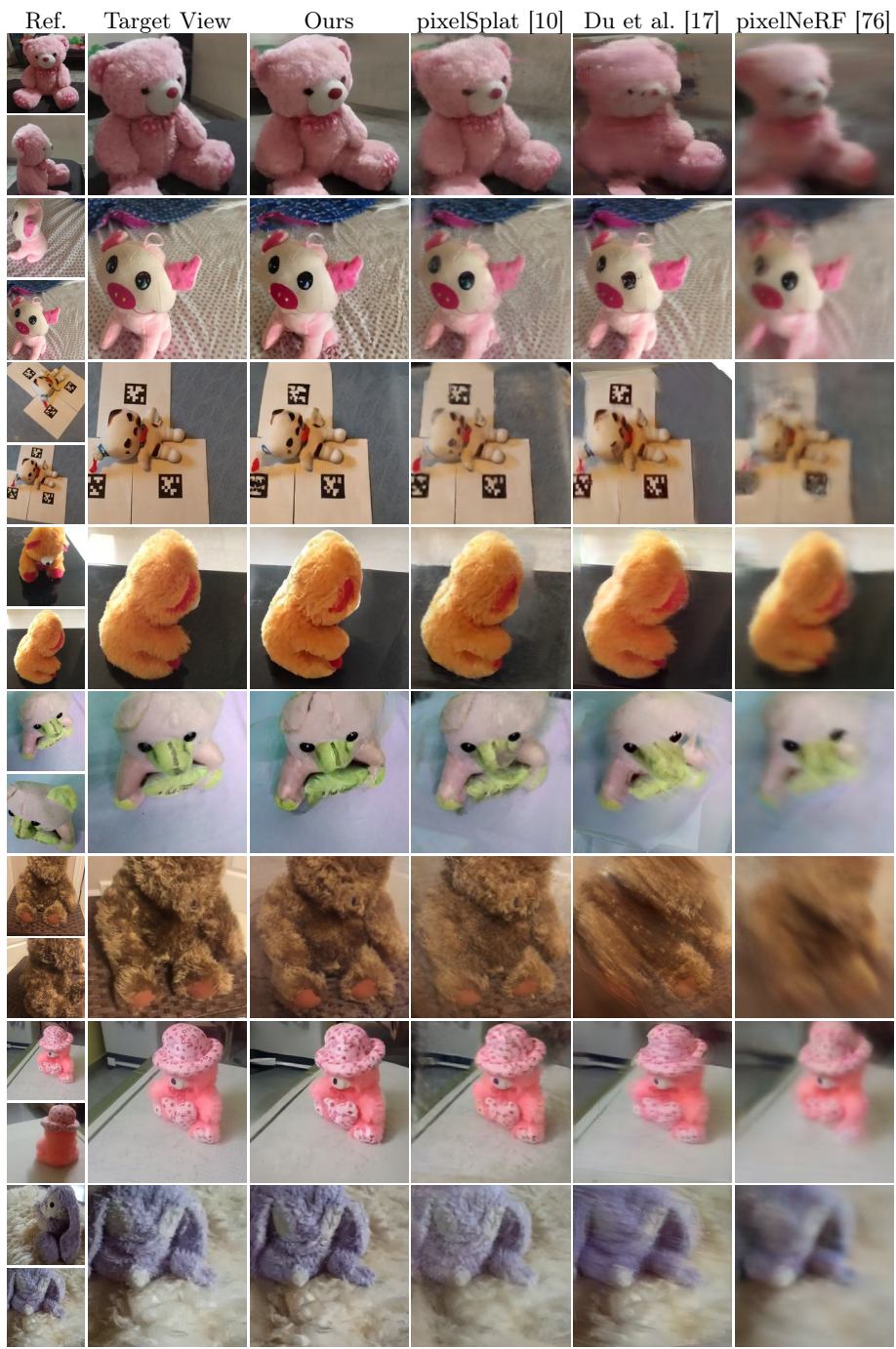}
    \caption{\textbf{Qualitative results for interpolation on CO3D teddybears~\cite{co3d}.}}
    \label{fig:qualitative_co3d_teddybear_intra}
\end{figure}

\begin{figure}[!t]
    \centering
    \includesvg[width = 1.06\textwidth]{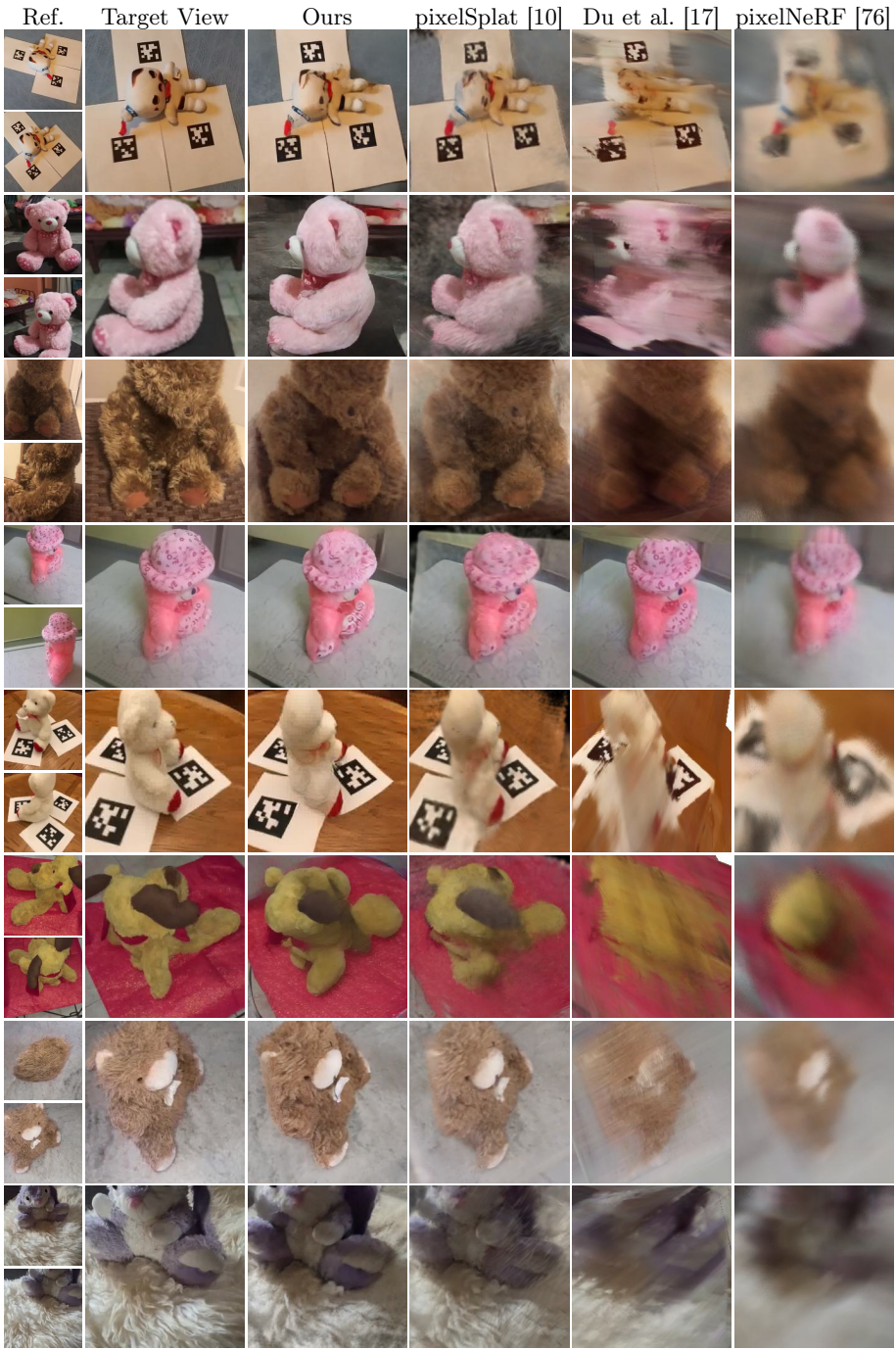}
    \caption{\textbf{Qualitative results for extrapolation on CO3D teddybears~\cite{co3d}.}}
    \label{fig:qualitative_co3d_teddybear_extra}
\end{figure}

\begin{figure}[!t]
    \centering
    \includesvg[width = 1.06\textwidth]{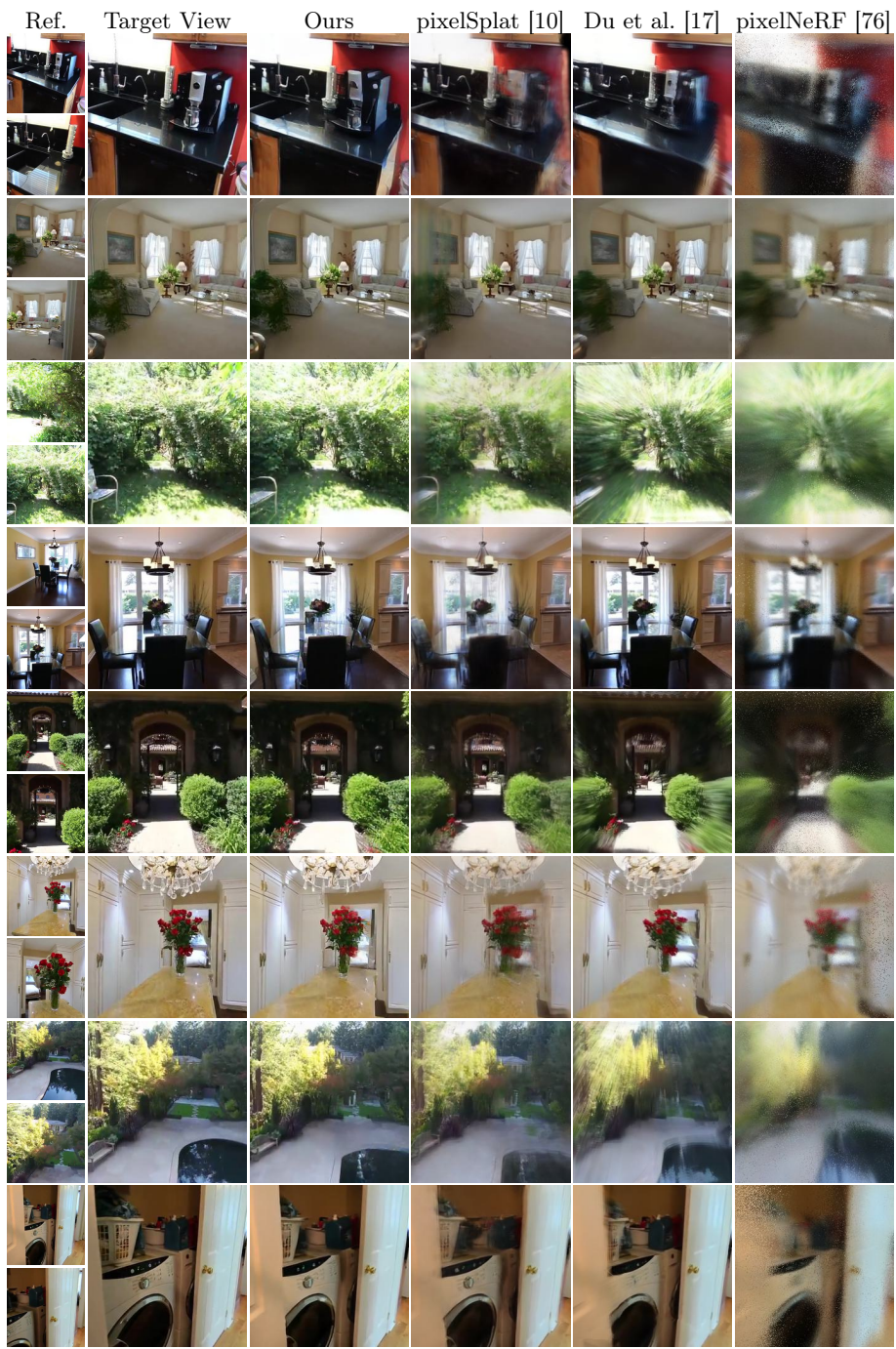}
    \caption{\textbf{Qualitative results for interpolation on RealEstate10k~\cite{Zhou2018multiplane}.}}
    \label{fig:qualitative_re10k_intra}
\end{figure}

\begin{figure}[!t]
    \centering
    \includesvg[width = 1.06\textwidth]{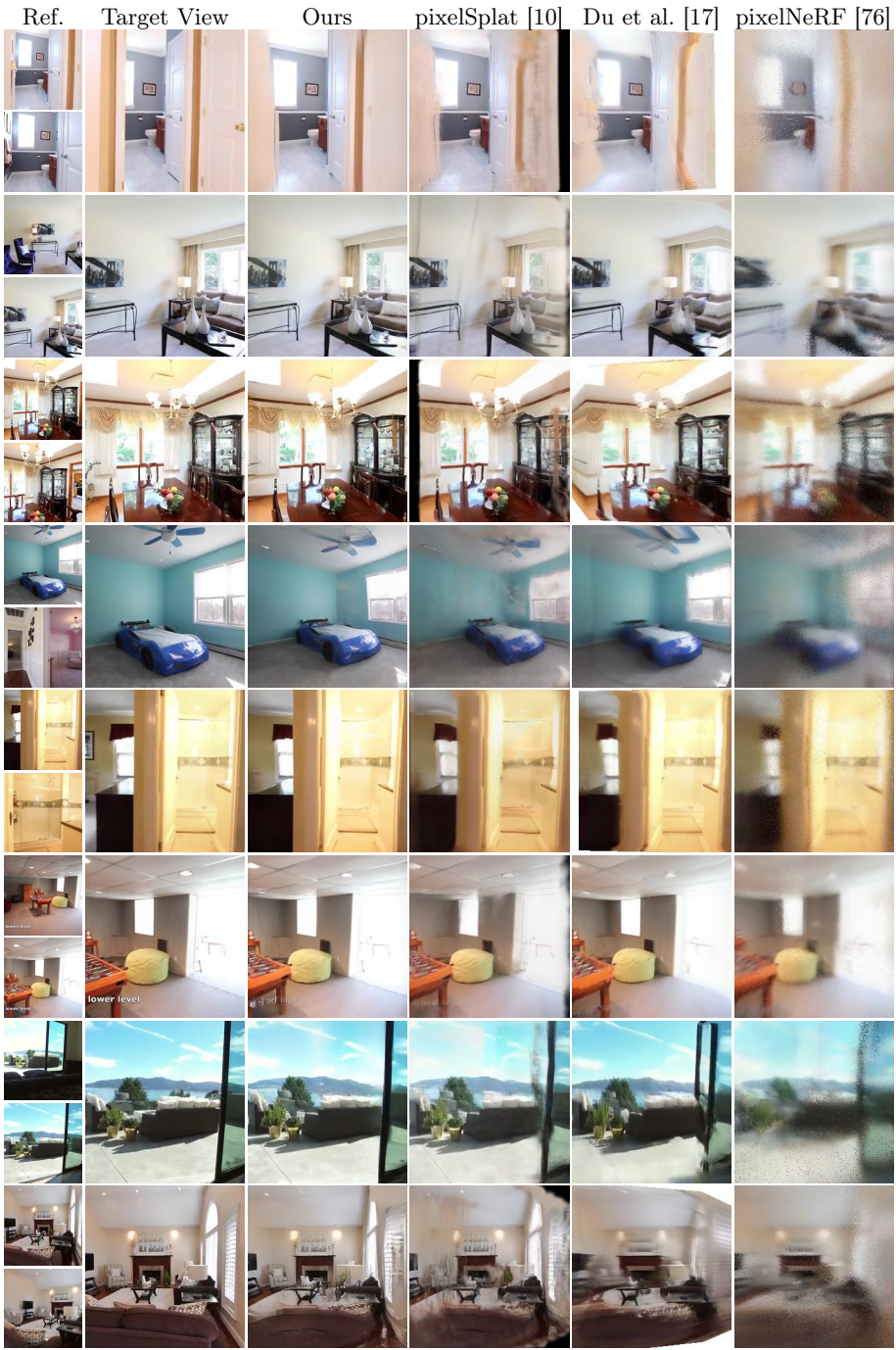}
    \caption{\textbf{Qualitative results for extrapolation on RealEstate10k~\cite{Zhou2018multiplane}.}}
    \label{fig:qualitative_re10k_extra}
\end{figure}

\end{document}